\documentclass[10pt, a4paper]{article}
\usepackage{lrec-coling2024} 

\usepackage{natbib}
\usepackage{multibib}
\usepackage{tikz-dependency}
\makeatletter
\def\@mb@citenamelist{cite,citep,citet,citealp,citealt,citepalias,citetalias}
\makeatother
\newcites{languageresource}{~}

\usepackage{graphicx}
\usepackage{tabularx}
\usepackage{soul}
\usepackage{titlesec}
\titleformat{\section}{\normalfont\large\bfseries\center}{\thesection.}{1em}{}
\titleformat{\subsection}{\normalfont\SmallTitleFont\bfseries\raggedright}{\thesubsection.}{1em}{}
\titleformat{\subsubsection}{\normalfont\normalsize\bfseries\raggedright}{\thesubsubsection.}{1em}{}
\renewcommand\thesection{\arabic{section}}
\renewcommand\thesubsection{\thesection.\arabic{subsection}}
\renewcommand\thesubsubsection{\thesubsection.\arabic{subsubsection}}

\usepackage{xcolor}
\usepackage{hyperref}
\definecolor{darkblue}{rgb}{0, 0, 0.5}
\hypersetup{colorlinks=true, citecolor=darkblue, linkcolor=darkblue, urlcolor=darkblue}

\usepackage{xstring}

\usepackage{color}

\usepackage{listings}

\title{Mathematical Entities: Corpora and Benchmarks}

\name{Jacob Collard, Valeria de Paiva, Eswaran Subrahmanian} 

\address{National Institute of Standards and Technology, Topos Institute, Carnegie Mellon University \\
         jacob.collard@nist.gov, valeria@topos.institute, es3e@andrew.cmu.edu}

\abstract{
Mathematics is a highly specialized domain with its own unique set of challenges. 
Despite this, there has been relatively little research on natural language processing for mathematical texts, and there are few mathematical language resources aimed at NLP. 
In this paper, we aim to provide annotated corpora that can be used to study the language of mathematics in different contexts, ranging from fundamental concepts found in textbooks to advanced research mathematics. 
We preprocess the corpora with a neural parsing model and some manual intervention to provide part-of-speech tags, lemmas, and dependency trees.
In total, we provide 182397 sentences across three corpora.
We then aim to test and evaluate several noteworthy natural language processing models using these corpora, to show how well they can adapt to the domain of mathematics and provide useful tools for exploring mathematical language.
We evaluate several neural and symbolic models against benchmarks that we extract from the corpus metadata to show that terminology extraction and definition extraction do not easily generalize to mathematics, and that additional work is needed to achieve good performance on these metrics.
Finally, we provide a learning assistant that grants access to the content of these corpora in a context-sensitive manner, utilizing text search and entity linking. 
Though our corpora and benchmarks provide useful metrics for evaluating mathematical language processing, further work is necessary to adapt models to mathematics in order to provide more effective learning assistants and apply NLP methods to different mathematical domains.
\\ \newline \Keywords{terminology extraction, definition extraction, entity linking, mathematics, category theory, information retrieval} }

\begin{document}

\maketitleabstract

\section{Introduction}
\label{sec:introduction}

The domain of mathematics has a number of unique features from the perspective of computational linguistics research.
Like most specialized domains, mathematics has its own vocabulary and quirks of language usage that differentiate it from other areas.
However, mathematical language also frequently contains inline formulas (where mathematical expressions are embedded in natural language), rigorously defined concepts, and formal language for describing proofs, theorems, and other mathematically rigorous statements. 
Mathematics is also an extremely multidisciplinary domain.
Different forms of mathematics are used in a wide variety of scientific domains.
Often, new branches of mathematics are applied to different domains, resulting in new areas of applied mathematics.
As a result, mathematical language can change quickly, and new, blended domains can arise mixing the language use of research mathematics and other scientific domains.
Finally, there is a limited amount of annotated data for mathematical language. 

These unique features grant particular importance to computational linguistic research on mathematics.
Any work on mathematical language will be applicable to a wide variety of scientific domains, and will improve the experience of researchers and students attempting to bridge gaps between their own fields of study and different sub-fields in mathematics. 
However, natural language processing for mathematics comes with its own unique set of challenges, as well. 
Processing formulas (or even language that simply contains formulas) can be especially difficult, as can reconciling the differences between everyday language and specialized mathematical terms.
Unfortunately, while there has been some recent work on natural language processing for mathematics, there is still a lack of benchmarks and comprehensive studies describing what is needed to fully take advantage of this fundamental domain.

Several tasks common to computational linguistics research are of special interest in mathematics:

\begin{itemize}
    \item Terminology extraction (TE) for identifying fundamental mathematical concepts themselves;
    \item Definition extraction (DE) for identifying formal definitions of mathematical concepts;
    \item Entity linking (EL) for connecting mathematical concepts to databases; and
    \item Collocation retrieval (CR) for identifying contexts of use for mathematical concepts.
\end{itemize}

There are other tasks that are also of interest that play a less significant role in this paper.
These include (but are not limited to) syntactic and semantic parsing of mathematical text, relation extraction, and extracting and linking other specialized environments such as proofs and theorems.
It is also possible to link natural language to formal systems such as theorem provers (e.g., Lean\footnote{\url{https://leanprover-community.github.io/}}, Isabelle\footnote{\url{ https://isabelle.in.tum.de/}}, and Coq\footnote{\url{https://coq.inria.fr/}}) or computer algebra systems (e.g., Sage\footnote{\url{https://www.sagemath.org/}} and GAP \footnote{\url{https://www.gap-system.org/}}).
Linking 
mathematics concepts both to a structured database representation and to proof assistants snippets of code is a newer task, \citetlanguageresource{MathGloss}.
These are important tasks, and we hope to provide some insight into them, but they are not the focus of this paper. 

Instead, our hope is to provide a collection of mathematical language corpora that provide insight and benchmarking for computational linguistics research on mathematics. 
This collection currently consists of three corpora, each of which covers a different context in the field of category theory. 
For these corpora, we provide benchmarks and discussions of several high-end models for terminology extraction and definition extraction, provide a simple model of entity linking, and show how an interface for collocation retrieval can aid in the use of these corpora as a resource for students and researchers. 
The benchmarks and corpora are available at \url{https://github.com/ToposInstitute/parmesan_benchmarks} and the learning assistant is available at \url{https://github.com/ToposInstitute/parmesan}.

\section{Previous Work}
\label{sec:previous-work}

\subsection{NLP and Mathematics}

There has been some scientific work on natural language processing for mathematical texts, sometimes referred to as mathematical language processing (MathLP). 
Most of these works focus on the representation and processing of mathematical formulas.
For example, \citet{kristiano2017} and \citet{Dadure2022} provide methods for representing mathematical formula for information retrieval.

\subsection{Terminology Extraction}
\label{subsec:previous-terminology-extraction}

Terminology extraction is the task of identifying the set of phrases in a text which represent key concepts in a domain.
Terminology extraction is closely related to named entity recognition, and often uses the same techniques; the difference is that terminology extraction is interested in a different set of entities, which usually are not people, places, or organizations. 
This is an important task for mathematics, since it can be used to identify key vocabulary for downstream tasks such as definition extraction and entity linking, and for the creation of indices or glossaries.

There has been a wide variety of research on terminology extraction.
Early methods made use of regular expressions or other rule-based operations applied to words or part-of-speech tags. 
The mwetoolkit3 \citelanguageresource{ramisch2012mwe}, for example, provides a framework for developing and searching regular expressions at different layers of representation to extract multi-word expressions, which may be candidates for terminology extraction.
TextRank \cite{mihalcea2004} is another early terminology extraction model that incorporates additional statistical information into a graph-based algorithm. 

More recent terminology extraction methods use deep learning. DyGIE++ \cite{Wadden2019EntityRA} combines entity extraction with relation and event extraction to achieve strong results on several benchmarks. 
It was followed by models such as SpERT.PL \cite{saietal2021spert} and PL-Marker \cite{ye2022plmarker}, which introduced additional linguistic information and a novel packing strategy, respectively.

Notably, many of these methods combine terminology extraction with relation extraction. 
Though relation extraction is not a primary goal of this paper, it is of interest to mathematics, since the relationships between mathematical concepts can be quite complex, but should be relatively well-defined.

Most previous work on terminology extraction has not been applied specifically to math.
However, many of the models mentioned above have been applied to other scientific domains and evaluated against the SciERC dataset \cite{luan2018multitask}, which consists of 500 scientific abstracts.
Since the sciences often make use of mathematical notation and concepts, it seems reasonable to expect that some transfer between the domains is possible. 

\subsection{Definition Extraction}
\label{subsec:previous-definition-extraction}

Definition extraction is the task of identifying the parts of a text that define a particular word or phrase.
There are often two components or strategies involved in definition extraction: identifying sentences that contain definitions, and identifying the terms and definitions precisely in text. Some models, such as \citep{ben_veyseh_joint_2019}, use a joint model which simultaneously identifies definitional sentences and precise terms and definitions.
This model is evaluated against three datasets: WCL, WC00, and DEFT. 
Word Class Lattices (WCL) is a definition extraction benchmark consisting of sentences from Wikipedia which distinguishes between definitional and non-definitional sentences \citep{navigli-velardi-2010-learning}.
WC00 also distinguishes between definitional and non-definitional sentences, and contains over 2000 sentences from the ACL anthology from the scientific domain \citep{jin-etal-2013-mining}.
DEFT consists of two subcorpora: one covering textbooks from domains including biology, history, and physics; and one covering contracts \citeplanguageresource{spala-etal-2019-deft}. 

The joint model described in \citet{ben_veyseh_joint_2019} reports an $F_1$ score of up to 85.3 on WCL, 66.9 on W00, 54.0 on DEFT Textbooks, and 71.7 on DEFT Contracts.
Another model, \citep{vanetik-etal-2020-automated} is designed specifically for mathematics and combines dependency tree and word vector representations to construct a neural classification model.
Their model scores above 0.9 on WCL and above 0.82 on WC00, and above 0.8 on WFM, a benchmark drawn from Wolfram MathWorld specifically for mathematics \citeplanguageresource{vanetik-etal-2019-wfm}.
This shows significant variation between different domains.
Ideally, we expect to have similar results for the category theory domain, but due to differences in the data that may not be reflected in training, some differences are possible.

\subsection{Entity Linking}
\label{subsec:previous-entity-linking}

Entity linking is the task of connecting an entity (often one extracted by a terminology extraction system) to a representation of that entity in a knowledge base such as WikiData\footnote{\url{wikidata.org}}.
Linking is similar in some ways to word sense disambiguation, in that the correct knowledge base record must be identified in the case that a word or phrase cannot be unambiguously attached to a single record. 
By linking entities to a knowledge base, a system can provide information about entities in a corpus from a manually curated repository such as WikiData.

As with terminology extraction, most entity linking methods have not been specifically evaluated on mathematical corpora. 
However, there have been a variety of entity linking models that have achieved good results in other areas.
\citet{raiman2018} uses a type system combined with a neural classifier to constrain and classify the entities associated with a candidate term. 
It is also common for entity linking models to make use of a knowledge graph, as \citet{mulang2020} do. 

\section{Corpus Development}
\label{sec:corpus-development}

We prepare three initial mathematical corpora for use in the study of mathematical language processing.
The first corpus consists of 755 abstracts (3188 sentences) from \emph{Theory and Applications of Categories} (TAC), a journal of category theory.
This corpus is very similar to the one presented in \citet{collard-etal-2022-extracting}, but has undergone additional processing and cleaning, which we describe here.
This corpus was selected as an exemplar for state-of-the-art mathematical research.
The abstracts contain many novel concepts as well as advanced contexts of use for fundamental concepts.

The second corpus consists of 11653 articles (175151 sentences) from the online encyclopedic resource for Category Theory nLab\footnote{\url{http://ncatlab.org}}. 
This corpus has undergone similar preprocessing to TAC.
It was selected as an exemplar for fundamental concepts in category theory and as a comprehensive reference.
In addition to mapping concepts directly to nLab articles, it is also possible to see concepts used in the context of other articles. 
For example, in addition to the article on categories itself, the word ``category'' appears in many other contexts within nLab that can help to elucidate its meaning.
To prepare this corpus for use, we remove the Markdown markup, leaving only plain text. 
We have also filtered out documents describing books as well as meta-articles such as lists and categories.

The third corpus consists of the entire text (4058 sentences) of \emph{Basic Category Theory} (BCT) by Tom Leinster \cite{leinster2014}. 
This is an introductory textbook, intended for students without advanced degrees in mathematics. 
As such, it is an exemplar of introductory concepts in category theory, similar to nLab, though  more foundational.
The text of the book is freely available for editing at \url{https://arxiv.org/abs/1612.09375}.
We process the entire \LaTeX{} code of the textbook with LaTeXML\footnote{\url{https://dlmf.nist.gov/LaTeXML}} to convert it into plain text for processing.

To handle \LaTeX{} markup in all three corpora, we use the LaTeXML converter to identify mathematical expressions. 
Completely removing mathematical expressions could introduce problems; since we later apply parsing to the corpora, the gaps caused by removing inline math will produce ungrammatical sentences and thus invalid dependency trees.
However, in their raw form, mathematical expressions are represented using \LaTeX{}, which can be difficult to read, especially when used to represent complex formulas.
Therefore, we convert mathematical expressions into plain text phrases using LaTeXML. These approximate the original mathematical formulas, providing the parser with linguistic material free of markup. For example, the expression \verb|\mathbb{Z}^n| can be represented as simply \verb|Z^n|.

Though these corpora do not include large amounts of annotation, they are associated with some useful metadata.
The TAC corpus includes titles, authors, dates, and keywords selected by the authors to describe their abstracts. 
These keywords will be used as part of the evaluation in the next section.
The nLab corpus includes titles and dates, of which the titles are used as part of the evaluation in the next section.
The BCT corpus contains LaTeX{} markup describing theorems and definitions, which provide additional context and can be used to evaluate definition extraction.

\begin{figure}
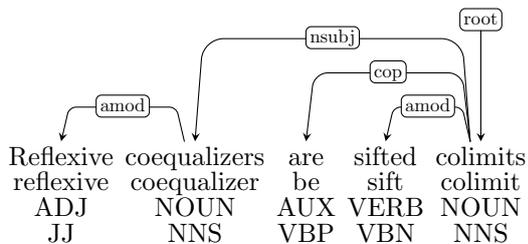

    \begin{dependency}
        \begin{deptext}
            Reflexive \& coequalizers \& are \& sifted \& colimits \\
            reflexive \& coequalizer \& be \& sift \& colimit \\
            ADJ \& NOUN \& AUX \& VERB \& NOUN \\
            JJ \& NNS \& VBP \& VBN \& NNS \\
        \end{deptext}
        \depedge{2}{1}{amod}
        \depedge{5}{2}{nsubj}
        \depedge{5}{3}{cop}
        \depedge{5}{4}{amod}
        \deproot{5}{root}
    \end{dependency}
    \caption{An annotated sentence from the TAC corpus with the original sentence, lemmas, coarse-grained and fine-grained POS tags, and labeled dependencies.\label{fig:annotated-dependency}}
\end{figure}

For all corpora, we also provide annotations of dependency trees, part of speech tags (using the universal dependencies tagset for coarse-grained annotations and spaCy's English tagset for fine-grained annotations), and lemmas in CONLL-U format\footnote{\url{https://universaldependencies.org/format.html}}. 
These annotations were generated automatically using the open NLP framework spaCy\footnote{\url{http://spacy.io}}.
The annotations are not fundamental to the rest of the work presented in this paper, and have not been rigorously evaluated.
However, we do hope to provide manual corrections and other improvements to the data so that these annotations can be used for evaluation.
An example annotated sentence is given in Figure \ref{fig:annotated-dependency}. 

\section{Experiments}
\label{sec:experiments}

We evaluate several high-performing models for each of three tasks: terminology extraction, definition extraction, and entity linking. 
Though we do not train these models, we hope to show how well current state-of-the-art models perform when generalizing to the domain of mathematics, and to highlight how our corpora can be used to evaluate these models. 

\subsection{Terminology Extraction}
\label{subsec:experiment-terminology-extraction}

We evaluate terminology extraction models against four benchmarks provided by our corpora: the set of author-selected keywords in TAC, the set of nLab titles, the set of glossary terms in BCT, and a set of automatically extracted multi-word expressions using mwetoolkit3. 
Alone, each of these evaluations is imperfect. 
There are many terms identified by each benchmark which are not identified by the others.
However, most of these benchmarks should exclusively contain valid mathematical concepts, with the exception of the automatically-extracted MWEs, which serve to identify novel concepts which none of the other benchmarks can. 

\begin{table}
    \centering
    \begin{tabular}{l|p{0.2\textwidth}}
        \textbf{Benchmark} & \textbf{Examples} \\
        \hline\hline
        Author Keywords & Abelian categorification, normal epimorphism, open map \\
        \hline
        nLab Titles & Balanced monoidal category, Fiber, Triple category \\
        \hline
        Glossary & Homotopy, Manifold, Partially ordered set \\
        \hline
        MWEs & Free double category, state sum construction, representable definition \\
    \end{tabular}
    \caption{Examples of each benchmark type}
\end{table}

To evaluate each model, we present it with the text of all three corpora and retrieve the set of extracted entities. 
This list is compared to the four benchmarks described above.
The models are not penalized for failing to extract specific instances of an entity; the only target is to extract the set of entities appearing in the corpus.

Table \ref{tab:terminology-extraction} shows the results of each model on each of the four benchmarks, as well as the combined score for all four benchmarks. 
The results are also broken down by corpus, since each corpus provides a different context of use which may be of interest to evaluation.

\begin{table*}[ht]
\begin{center}
\small
\begin{tabular}{|l|lll|lll|lll|lll|lll|}
\hline
~ & \multicolumn{3}{c|}{\textbf{BCT Glossary}} & \multicolumn{3}{c|}{\textbf{Keywords}} & \multicolumn{3}{c|}{\textbf{Titles}} & \multicolumn{3}{c|}{\textbf{MWEs}} & \multicolumn{3}{c|}{\textbf{Combined}} \\
\hline
~ & P & R & F1 & P & R & F1 & P & R & F1 & P & R & F1 & P & R & F1 \\
\hline\hline
\multicolumn{16}{|l|}{\textbf{TAC Corpus}} \\
\hline\hline
Textrank & 0.13 & 0.60 & 0.21 & 0.15 & 0.55 & 0.23 & 0.09 & 0.46 & 0.14 & 0.25 & 0.78 & 0.38 & 0.15 & 0.59 & 0.24 \\
DyGIE++ & 0.18 & 0.38 & 0.24 & 0.22 & 0.35 & 0.27 & 0.12 & 0.27 & 0.16 & 0.28 & 0.70 & 0.40 & 0.2 & 0.43 & 0.27 \\
SpERT.PL & 0.10 & 0.66 & 0.17 & 0.14 & 0.77 & 0.23 & 0.08 & 0.63 & 0.14 & 0.33 & 0.68 & 0.44 & 0.16 & 0.69 & 0.26 \\
PL-Marker & 0.22 & 0.40 & 0.28 & 0.23 & 0.38 & 0.28 & 0.11 & 0.27 & 0.16 & 0.30 & 0.60 & 0.4 & 0.21 & 0.41 & 0.28 \\
\hline\hline
\multicolumn{16}{|l|}{\textbf{nLab Corpus}} \\
\hline\hline
Textrank & 0.08 & 0.68 & 0.14 & 0.12 & 0.65 & 0.23 & 0.08 & 0.55 & 0.14 & 0.21 & 0.69 & 0.32 & 0.12 & 0.64 & 0.20 \\
DyGIE++ & 0.14 & 0.58 & 0.23 & 0.20 & 0.46 & 0.28 & 0.04 & 0.60 & 0.08 & 0.25 & 0.72 & 0.37 & 0.16 & 0.50 & 0.24 \\
SpERT.PL & 0.05 & 0.67 & 0.09 & 0.09 & 0.65 & 0.16 & 0.03 & 0.68 & 0.06 & 0.22 & 0.72 & 0.34 & 0.12 & 0.65 & 0.20 \\
PL-Marker & 0.34 & 0.66 & 0.45 & 0.25 & 0.44 & 0.32 & 0.15 & 0.34 & 0.21 & 0.35
& 0.65 & 0.45 & 0.28 & 0.50 & 0.36 \\
\hline\hline
\multicolumn{16}{|l|}{\textbf{BCT Corpus}} \\
\hline\hline
Textrank & 0.36 & 0.81 & 0.50 & 0.29 & 0.70 & 0.41 & 0.32 & 0.82 & 0.46 & 0.48 & 0.88 & 0.62 & 0.40 & 0.64 & 0.49 \\
DyGIE++ & 0.23 & 0.44 & 0.30 & 0.34 & 0.66 & 0.44 & 0.22 & 0.55 & 0.31 & 0.34 & 0.73 & 0.46 & 0.28 & 0.60 & 0.38 \\
SpERT.PL & 0.20 & 0.68 & 0.3 & 0.18 & 0.81 & 0.30 & 0.15 & 0.68 & 0.25 & 0.47 & 0.77 & 0.58 & 0.25 & 0.73 & 0.37 \\
PL-Marker & 0.35 & 0.61 & 0.44 & 0.31 & 0.52 & 0.39 & 0.29 & 0.52 & 0.37 & 0.43 & 0.81 & 0.56 & 0.36 & 0.63 & 0.46 \\
\hline\hline
\multicolumn{16}{|l|}{\textbf{Combined Corpus}} \\
\hline\hline
Textrank & 0.19 & 0.70 & 0.30 & 0.19 & 0.63 & 0.29 & 0.16 & 0.61 & 0.25 & 0.22 & 0.31 & 0.26 & 0.19 & 0.56 & 0.28 \\
DyGIE++ & 0.18 & 0.47 & 0.26 & 0.13 & 0.49 & 0.21 & 0.13 & 0.47 & 0.20 & 0.21 & 0.29 & 0.24 & 0.16 & 0.43 & 0.23 \\
SpERT.PL & 0.12 & 0.67 & 0.20 & 0.14 & 0.74 & 0.24 & 0.09 & 0.66 & 0.16 & 0.18 & 0.34 & 0.24 & 0.13 & 0.60 & 0.21 \\
PL-Marker & 0.30 & 0.56 & 0.39 & 0.26 & 0.45 & 0.33 & 0.18 & 0.38 & 0.24 & 0.28 & 0.36 & 0.32 & 0.26 & 0.44 & 0.33 \\
\hline
\end{tabular}
\caption{Results of terminology extraction on mathematical corpora. Each column represents an evaluation benchmark, while each row represents a model applied to a particular corpus.\label{tab:terminology-extraction}}
\end{center}
\end{table*}

\subsection{Definition Extraction}
\label{subsec:experiment-definition-extraction}

To provide a benchmark for definition extraction, we use the definition environments found in BCT. 
These definitions were explicitly identified by the author in the LaTeX{} code, though they may not be the only definitional statements in the book. 
Each of the evaluated models is given the entire text of the book, with the task of identifying definitional content as well as the headword for each definition. 
We record precision, recall, and F1 scores for the number of words matched between the benchmark definition and the predicted definition.
Table \ref{tab:definition-extraction} shows the results for two advanced definition extraction systems.

\begin{table}[ht]
\begin{center}
\begin{tabular}{|l|lll|}
\hline
~ & \textbf{Precision} & \textbf{Recall} & \textbf{F1} \\
\hline
\textbf{Vanetik et al.} & 0.12 & 0.44 & 0.19 \\
\textbf{Veyseh et al.} & 0.03 & 0.33 & 0.05 \\
\hline
\end{tabular}
\caption{Results of definition extraction on BCT\label{tab:definition-extraction}}
\end{center}
\end{table}

\subsection{Entity Linking}
\label{subsec:experiment-entity-linking}

To evaluate entity linking, we have constructed a set of 126 distinct mathematical concepts and identified WikiData entries that correspond to them within the field of category theory.
This correspondence was determined manually by a mathematician. 
In some cases, there are still multiple WikiData entries that could correspond to the entity. 
In these cases, all possible entries are included.

We evaluate two entity linking models: a simple query-based model using the Wikidata query service\footnote{\url{https://query.wikidata.org/}} and a simple neural model\footnote{\url{https://github.com/egerber/spaCy-entity-linker/tree/master/spacy_entity_linker}}.

We developed the query-based model as a simple way to retrieve Wikidata entries that are likely to be in the field of category theory, as opposed to other domains. 
The complete query we used is provided in the supplementary code.
This query finds entries whose label or alias matches the given phrase, but filters out any entries which belong to the following classes, which are unlikely to contain mathematical concepts: physical objects, concrete objects, physical locations, Wikimedia categories, activities, human behaviors, artistic concepts, points in time, time intervals, and currencies.


The query may seem somewhat arbitrary; it was developed over time during the course of this project to remove specific errors that we found. 
We expect to continue developing this query to better match the needs of users and improve the results of the evaluation.

Table \ref{tab:entity-linking} shows the results of two entity linking models on our benchmark. 
We provide precision at 1 (P@1), recall, and $F_1$ score. 
P@1 is used since the entity linking model may provide many potential candidates in a ranked list, the latter of which are much less likely to be valid.

\begin{table}[ht]
\begin{center}
\begin{tabular}{|l|lll|}
\hline
~ & \textbf{P@1} & \textbf{Recall} & \textbf{F1} \\
\hline
\textbf{Query} & 0.60 & 0.82 & 0.68 \\
\textbf{spaCy} & 0.59 & 0.52 & 0.55 \\
\hline
\end{tabular}
\caption{Results of entity linking for category theory.\label{tab:entity-linking}}
\end{center}
\end{table}

\section{Learning Assistant}
\label{sec:search-retrieval}

\begin{figure*}[!ht]
\begin{center}

\includegraphics[width=\textwidth]{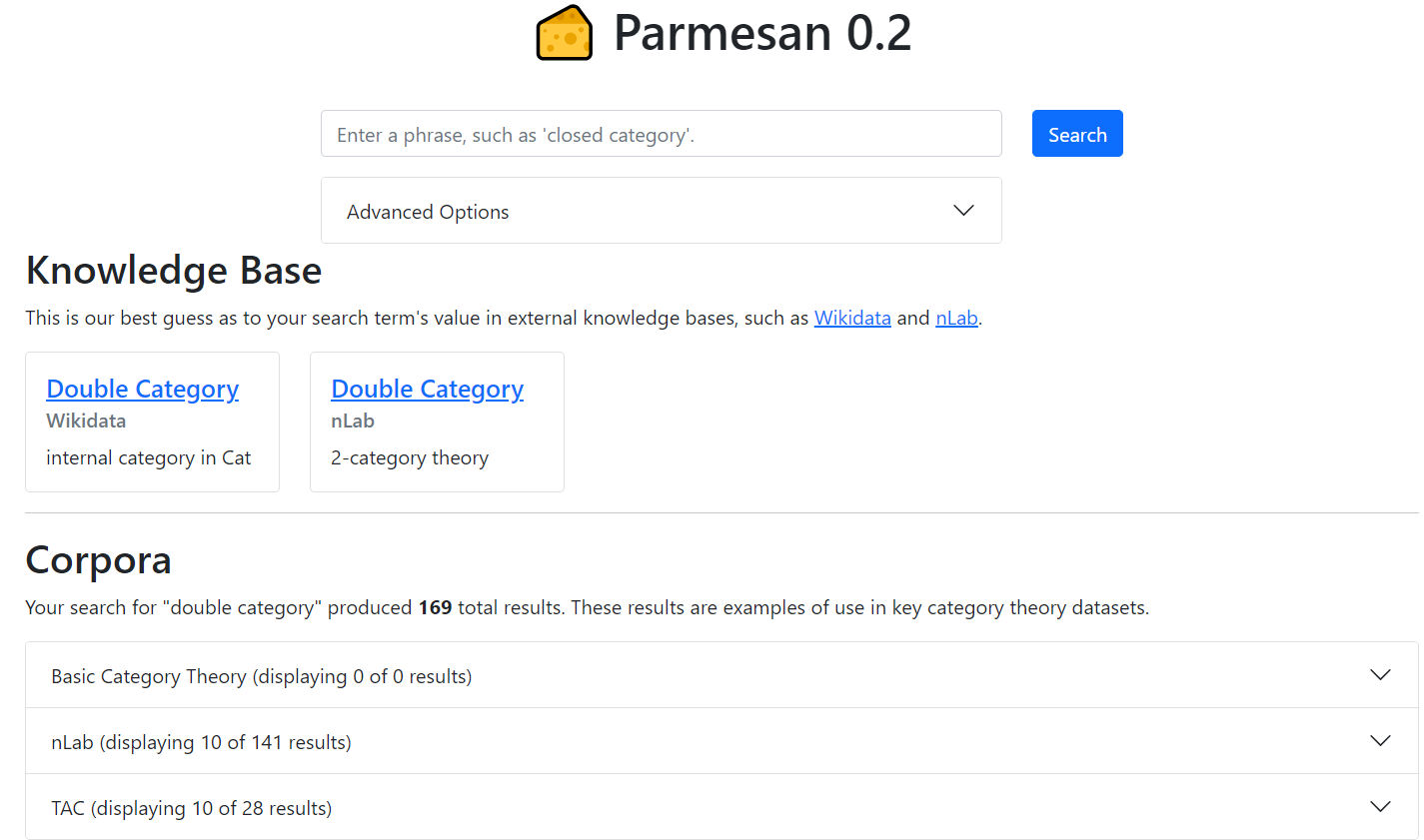}
\includegraphics[width=\textwidth]
{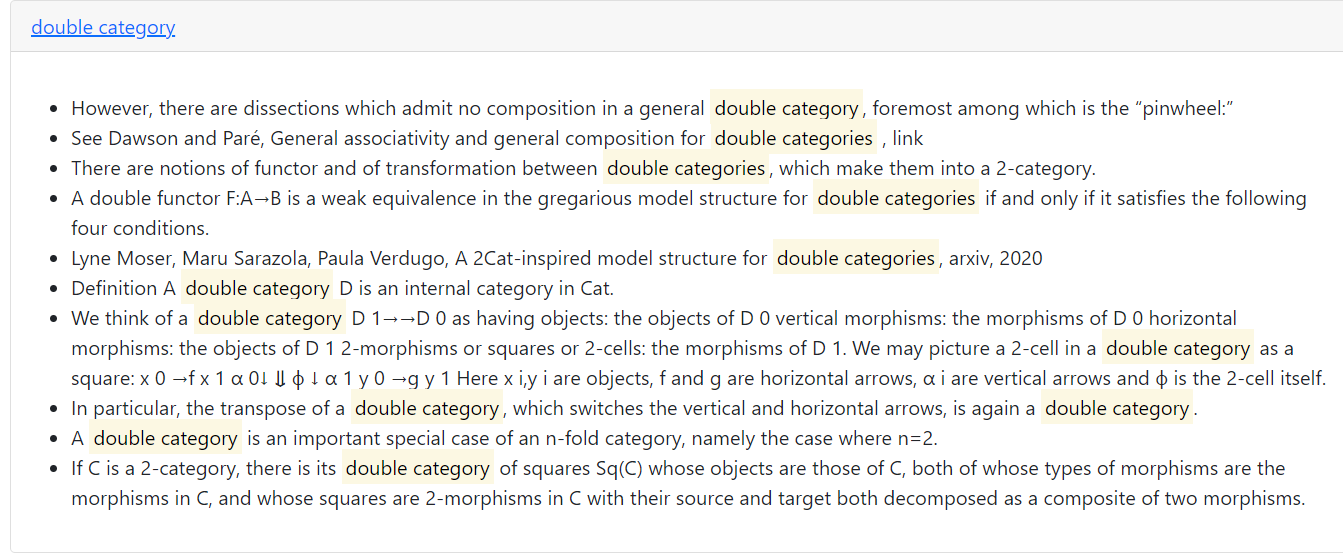}
\includegraphics[width=\textwidth]
{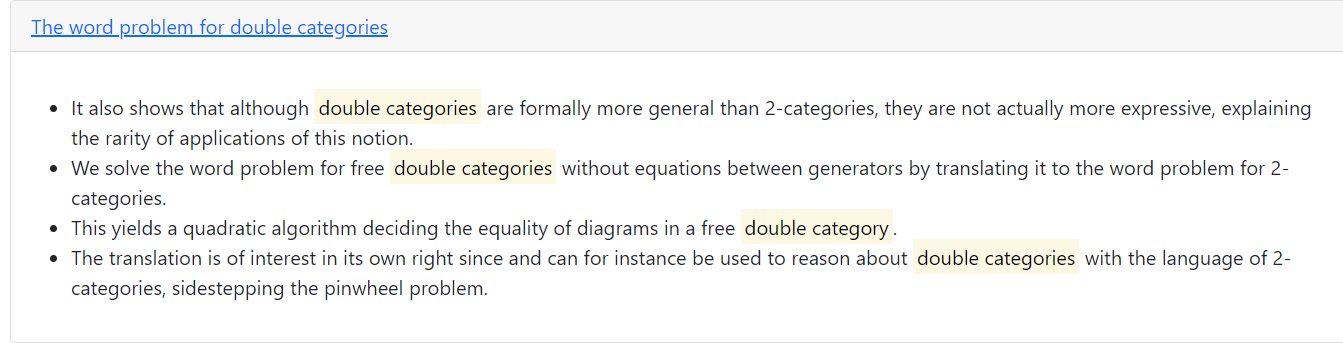}

\caption{Search results using our user interface. The given results are from nLab and TAC, respectively.}
\label{fig:search-results}
\end{center}
\end{figure*}

In addition to the corpora and benchmarks described above, we have developed a simple learning assistant called Parmesan (PARsing Mathematical Entities Search And Navigation) 
that provides text search over data in our corpora.
The input to the interface is a term, which may be any word or phrase that appears in the corpora, and the output is a set of sentences in which the term occurs, as well as links to Wikidata and nLab entries.
This search is intended to allow users to identify the contexts in which an unfamiliar term appears. 
This complements the definitions provided by entity linking by showing how the terms are actually used in different mathematical contexts.

There are two types of context that the learning assistant provides.
At a high level, the three corpora (TAC, nLab, and BCT) represent contexts of language use.
The TAC corpus, being drawn from journal articles, provides a view into the state-of-the-art, advanced concepts, and newly-coined phrases.
The nLab and BCT corpora, on the other hand, are primarily dedicated to descriptions of common, high-level concepts in category theory.

Each corpus also provide precise contexts in the form of exemplars of sentences and phrases where the target term is used.
We can easily identify matches of the phrase that the user has input by finding corresponding lemmas in the annotated corpus.
We use lemmas (generated with spaCy) to show the term used with different inflections.
We then display all sentences that contain the corresponding lemmas. 
However, the distinction between the three corpora is kept clear: results from TAC are returned separately from nLab, which are returned separately from BCT. 
This allows the user to clearly distinguish between different contexts in which terms appear, both at the sentential level, at the document level, and at the corpus level.
Links are provided to individual nLab articles and to specific TAC abstracts if the user requires more information or additional context.

Figure \ref{fig:search-results} shows an example of search results found by the search engine for the search term ``double category''.
At the top is the list of knowledge base entries found for that concept; in this case, there is  the concept of a double category from category theory in WikiData (Wikidata entry Q99613675) and the nLab entry on double categories. 

Next are shown results from BCT, nLab, and TAC sentences.
Each document is displayed as a separate card with a link to the original document (TAC abstract, nLab article, or BCT paragraph). 
Notably, there are no results for double category in BCT, indicating that it is a slightly more advanced term not found in a typical introductory course. 
A list of sentences containing the search term are then shown within the card.
The search term is highlighted where it appears in the text of each sentence.
As can be seen by this example, variant forms of the word (such as the plural ``categories'') are shown as well as the exact terms searched by the user.
However, there is currently no additional semantic or vector-based search to identify similar concepts to ``double'' or ``category''. 
Since the system is aimed primarily at learners, we hope to keep the analysis straightforward and easily interpretable to the user. 

The example sentences in Figure \ref{fig:search-results} reveal certain facts about double categories that are useful to a newcomer in the field: they are formally more general than 2-categories; there is a kind of double category called a free double category; there are certain mathematical problems of interest for double categories.


The user is also able to hide and display the TAC and/or nLab and BCT corpora individually if they are only searching for information from a certain set of contexts.

\section{Discussion}
\label{sec:discussion}

\subsection{Mathematical Language Processing}
\label{subsec:mathematical-language-processing}

The experiments in Section \ref{sec:experiments} show that additional work is necessary to achieve strong performance in mathematical language processing. 
Though SpERT.PL achieves high recall, this is coupled with low precision, suggesting that many of the terms this model predicts are not actually valid mathematical terms.
This can be confirmed qualitatively by examining the set of false positives found by each model, as shown in Table \ref{tab:false-positives}.
Many of these predicted terms, though valid phrases, are not mathematical in nature and do not refer to specific concepts.

\begin{table}[ht]
\begin{center}
\begin{tabular}{|l|p{0.6\linewidth}|}
\hline
\textbf{DyGIE++} & all, also to obtain, be necessarily, e / m, if \\
\hline
\textbf{Textrank} & 1965, all, any diagram, at least one, several approaches \\
\hline
\textbf{SpERT.PL} & it, them, basis, R(S), one \\
\hline
\textbf{PL-Marker} & and both, these, ideas, to, a choice of a \\
\hline
\end{tabular}
\caption{Example false positives for terminology extraction.\label{tab:false-positives}}
\end{center}
\end{table}

These models were not given the opportunity to adapt to the category theory domain through training.
Providing some training examples from category theory has the potential to improve these results. 
However, it should also be noted that Textrank is an unsupervised model of terminology extraction, and still underperforms on mathematical language relative to its performance in other domains \cite{mihalcea2004}. 
This may suggest that there is inherent difficulty identifying mathematical concepts. 

There are similar challenges for definition extraction, and we can likewise confirm through false positives that many of the results are unexpected for mathematics, as shown in Tables \ref{tab:joint-model-definitions} and \ref{tab:vanetik-model-definitions}. 

\begin{table*}
    \centering
    \begin{tabular}{|l|p{22em}|}
        \hline
        Term & Definition \\
        \hline
        composition & describes the process of attaching the outputs of one circuit to the inputs of another \\
        \hline
        a decomposition & the composition-representative subsets of the hom-set $T([m],[0$ \\
        \hline
        isotropy group & a presheaf of groups on C \\
        \hline
        distributive & finite \\
        \hline
        distributive & both a tensor and a par \\
        \hline
        idempotent relations & the \\
        \hline
        cartesian & every comonad has an Eilenberg-Moore object and every left adjoint arrow is \\
        \hline
        isotropy rank & the isotropy rank of a small category is the ordinal at which the sequence of quotients stabilizes \\
        \hline
        FILL & an intriguing version of \\
        \hline
        Cat & Lax$_N(B$ \\
        \hline
    \end{tabular}
    \caption{Definitions extracted by \citet{ben_veyseh_joint_2019}}
    \label{tab:joint-model-definitions}
\end{table*}

\begin{table*}
    \centering
    \begin{tabular}{|p{27em}|}
        \hline
        The equivalence is FOLDS equivalence of the FOLDS-Specifications of the two concepts. \\
        \hline
        The concept of algebra is given as an adjunction with invertible counit. \\
        \hline
        Thus we maintain that the notion of linear-distributive category (which has both a tensor and a par, but is nevertheless more general than the notion of monoidal category) provides the correct framework in which to interpret the concept of Frobenius algebra. \\
        \hline
        The goal of this article is to emphasize the role of cubical sets in enriched category theory and infinity-category theory. \\
        \hline
        A model for an EA sketch in a lextensive category is a ‘snapshot’ of a database with values in that category. \\
        \hline
    \end{tabular}
    \caption{Definitional sentences identified by \citet{vanetik-etal-2020-automated}}
    \label{tab:vanetik-model-definitions}
\end{table*}

The entity linking models we present, including the simple query-based model, perform relatively well, however, possibly due to a relatively low incidence of ambiguity in category theory.
Though some specific terms are highly ambiguous, other mathematical concepts are complex phrases, which do not have everyday meanings.
Additional stress-testing with sets of shorter phrases may reveal additional challenges in the area.

\subsection{Mathematical Information Retrieval}
\label{subsec:mathematical-information-retrieval}

The current implementation of the user interface provides a tool for learners and researchers in the field of category theory to search for concepts to find their contexts of use and information about them in Wikidata.
This provides the user with different points of view about a concept: concise but highly structured, interconnected data in Wikidata; the expert, but general and pedagogical, view of nLab; and the cutting-edge research point-of-view in TAC.

Each of these points of view may be useful to different users, and separating them in the display allows the user to compare and contrast different contexts of use of the words they are looking for, providing real-world examples and practical information about novel concepts.

This style of interactive learning can be further improved as we incorporate resources from other sources and new natural language processing methods.
For example, new corpora can be incorporated to provide new contexts of use for concepts. 
Adding a repository of articles from a category theory subsection of arXiv would add contexts from new preprints and a broader class of mathematical journal articles.
Similarly, we can incorporate entity linking to other databases such as Planet Math\footnote{\url{https://planetmath.org}} or the Encyclopedia of Mathematics\footnote{\url{https://encyclopediaofmath.org/wiki/Main_Page}}.

We can also incorporate new advances in natural language processing and technology.
As shown in Section \ref{sec:experiments}, terminology extraction suffers from challenges in specific domains such as category theory.
Since the relation extraction algorithms we study are unable to accurately extract mathematical concepts, the relations that build on these concepts are generally lacking as well.
With additional training or other advances in relation extraction, the addition of relations to the interface would introduce a new type of context to users.
By understanding how concepts are related to one another, a learner can understand the meaning of that concept in terms of more familiar ideas.
Adding definition extraction, semantic similarity search, and other natural language processing methods to the system can grant it similar improvements.

Other future work for the system includes improving the order of search results, better filters on Wikidata links, and various performance improvements.
The addition of automatic definition extraction is also considered to be of particular importance, since definitions as they appear in context will be especially useful to learners. 

The principles of this research are by no means limited to category theory, though category theory does pose some unique challenges and provides some unique opportunities due to its growing presence in interdisciplinary research.
Similar interfaces could, however, be applied to any field.

Overall, our work provides a new approach to search for learners new to the field of category theory.
This approach is centered around providing context and domain-specific knowledge about user concepts.
Because the user provides the concepts, there is less need for error-prone concept extraction, and we can instead rely on entity linking, taking advantage of known properties of the domain.
The system provides several different contexts, allowing the user to compare and contrast disparate sources of knowledge to find the information they need about novel concepts.

We have shown that state-of-the-art computational linguistic tools largely do not apply, without adaptation, to mathematical texts. 
Precision and recall scores are much lower than originally expected.
However, it may be possible to adapt these models more effectively with additional training provided by annotated corpora, vocabularies, and knowledge graphs. 
We have provided some initial linguistically annotated mathematical corpora and online tools to build up the toolbox for processing mathematical texts. 
Much more additional work is needed.
We hope to continue work in category theory, improving our corpora, building up a knowledge graph for category theory using definitions and relation identification, as well as representing mathematical results (theorems, lemmas, and propositions).
We have also begun extending this work into a corpus of linear algebra, showing that the results are not specific to category theory.
Our work complements previous work concentrating on proofs by targeting mathematical statements, definitions, and concepts.

\section{Disclaimer}

Certain commercial entities, equipment, or materials may be identified in this document in order to describe an experimental procedure or concept adequately.
Such identification is not intended to imply recommendation or endorsement by the National Institute of Standards and Technology, nor is it intended to imply that the entities, materials, or equipment are necessarily the best available for the purpose.

\section{Bibliographical References}\label{reference}

\bibliographystyle{lrec-coling2024-natbib}
\bibliography{lrec-coling2024-example}

\begin{thebibliography}{4}
\expandafter\ifx\csname natexlab\endcsname\relax\def\natexlab#1{#1}\fi

\bibitem[{Horowitz and de~Paiva(2023)}]{MathGloss}
Lucy Horowitz and Valeria de~Paiva. 2023.
\newblock \href {https://europroofnet.github.io/cambridge-2023/#horowitz}
  {Mathgloss: Linked undergraduate math concepts}.
\newblock EuroProofNet Workshop on Natural Formal Mathematics and Libraries of
  Formal Proofs and Natural Mathematical Language.

\bibitem[{{Ramisch}(2012)}]{ramisch2012mwe}
Carlos {Ramisch}. 2012.
\newblock A generic framework for multiword expressions treatment: From
  acquisition to applications.
\newblock In \emph{Proceedings of the {ACL} 2012 Student Research Workshop},
  Jeju, Republic of Korea.

\bibitem[{Spala et~al.(2019)Spala, Miller, Yang, Dernoncourt, and
  Dockhorn}]{spala-etal-2019-deft}
Sasha Spala, Nicholas~A. Miller, Yiming Yang, Franck Dernoncourt, and Carl
  Dockhorn. 2019.
\newblock \href {https://doi.org/10.18653/v1/W19-4015} {{DEFT}: A corpus for
  definition extraction in free- and semi-structured text}.
\newblock In \emph{Proceedings of the 13th Linguistic Annotation Workshop},
  pages 124--131, Florence, Italy. Association for Computational Linguistics.

\bibitem[{Vanetik et~al.(2019)Vanetik, Litvak, Shevchuk, and
  Reznik}]{vanetik-etal-2019-wfm}
Natalia Vanetik, Marina Litvak, Sergey Shevchuk, and Lior Reznik. 2019.
\newblock \href
  {https://github.com/uplink007/FinalProject/tree/master/data/wolfram} {{WFM}
  dataset of mathematical definitions}.

\end{thebibliography}


\begin{thebibliography}{15}
\expandafter\ifx\csname natexlab\endcsname\relax\def\natexlab#1{#1}\fi

\bibitem[{Collard et~al.(2022)Collard, de~Paiva, Fong, and
  Subrahmanian}]{collard-etal-2022-extracting}
Jacob Collard, Valeria de~Paiva, Brendan Fong, and Eswaran Subrahmanian. 2022.
\newblock \href {https://aclanthology.org/2022.wnut-1.2} {Extracting
  mathematical concepts from text}.
\newblock In \emph{Proceedings of the Eighth Workshop on Noisy User-generated
  Text (W-NUT 2022)}, pages 15--23, Gyeongju, Republic of Korea. Association
  for Computational Linguistics.

\bibitem[{Dadure et~al.(2022)Dadure, Pakray, and Bandyopadhyay}]{Dadure2022}
Pankaj Dadure, Partha Pakray, and Sivaji Bandyopadhyay. 2022.
\newblock \href {https://doi.org/https://doi.org/10.1016/j.jksuci.2021.05.014}
  {Embedding and generalization of formula with context in the retrieval of
  mathematical information}.
\newblock \emph{Journal of King Saud University - Computer and Information
  Sciences}, 34(9):6624--6634.

\bibitem[{Jin et~al.(2013)Jin, Kan, Ng, and He}]{jin-etal-2013-mining}
Yiping Jin, Min-Yen Kan, Jun-Ping Ng, and Xiangnan He. 2013.
\newblock \href {https://aclanthology.org/D13-1073} {Mining scientific terms
  and their definitions: A study of the {ACL} {A}nthology}.
\newblock In \emph{Proceedings of the 2013 Conference on Empirical Methods in
  Natural Language Processing}, pages 780--790, Seattle, Washington, USA.
  Association for Computational Linguistics.

\bibitem[{Kristianto et~al.(2017)Kristianto, Topić, and
  Aizawa}]{kristiano2017}
Giovanni~Yoko Kristianto, Goran Topić, and Akiko Aizawa. 2017.
\newblock \href {https://doi.org/https://doi.org/10.1007/s10791-017-9296-8}
  {Utilizing dependency relationships between math expressions in math {IR}}.
\newblock \emph{Information Retrieval Journal}, 20(2):132--167.

\bibitem[{{Leinster}(2014)}]{leinster2014}
Tom {Leinster}. 2014.
\newblock \emph{Basic Category Theory}.
\newblock Cambridge University Press.

\bibitem[{Luan et~al.(2018)Luan, He, Ostendorf, and
  Hajishirzi}]{luan2018multitask}
Yi~Luan, Luheng He, Mari Ostendorf, and Hannaneh Hajishirzi. 2018.
\newblock Multi-task identification of entities, relations, and coreferencefor
  scientific knowledge graph construction.
\newblock In \emph{Proc.\ Conf. Empirical Methods Natural Language Process.
  (EMNLP)}.

\bibitem[{Mihalcea and Tarau(2004)}]{mihalcea2004}
Rada Mihalcea and Paul Tarau. 2004.
\newblock \href {https://aclanthology.org/W04-3252} {{T}ext{R}ank: Bringing
  order into text}.
\newblock In \emph{Proceedings of the 2004 Conference on Empirical Methods in
  Natural Language Processing}, pages 404--411, Barcelona, Spain. Association
  for Computational Linguistics.

\bibitem[{Mulang' et~al.(2020)Mulang', Singh, Prabhu, Nadgeri, Hoffart, and
  Lehmann}]{mulang2020}
Isaiah~Onando Mulang', Kuldeep Singh, Chaitali Prabhu, Abhishek Nadgeri,
  Johannes Hoffart, and Jens Lehmann. 2020.
\newblock \href {https://doi.org/10.1145/3340531.3412159} {Evaluating the
  impact of knowledge graph context on entity disambiguation models}.
\newblock In \emph{Proceedings of the 29th ACM International Conference on
  Information \& Knowledge Management}, CIKM '20, page 2157–2160, New York,
  NY, USA. Association for Computing Machinery.

\bibitem[{Navigli and Velardi(2010)}]{navigli-velardi-2010-learning}
Roberto Navigli and Paola Velardi. 2010.
\newblock \href {https://aclanthology.org/P10-1134} {Learning word-class
  lattices for definition and hypernym extraction}.
\newblock In \emph{Proceedings of the 48th Annual Meeting of the Association
  for Computational Linguistics}, pages 1318--1327, Uppsala, Sweden.
  Association for Computational Linguistics.

\bibitem[{{Raiman} and {Raiman}(2018)}]{raiman2018}
Johnathan {Raiman} and Oliver {Raiman}. 2018.
\newblock \href {https://doi.org/10.1609/aaai.v32i1.12008} {Deeptype:
  Multilingual entity linking by neural type system evolution}.
\newblock In \emph{Proceedings of the Thirty-Second {AAAI} Conference on
  Artificial Intelligence ({AAAI}-18)}, pages 5406--5413.

\bibitem[{{Sai} et~al.(2021){Sai}, {Chakraborty}, {Dutta}, {Sanyal}, and
  {Das}}]{saietal2021spert}
Santosh Tokala Yaswanth~Sri {Sai}, Prantika {Chakraborty}, Sudakshina {Dutta},
  Debarshi~Kumar {Sanyal}, and Partha~Pratim {Das}. 2021.
\newblock Joint entity and relation extraction from scientific documents: Role
  of linguistic information and entity types.
\newblock In \emph{2nd Workshop on Extraction and Evaluation of Knowledge
  Entities from Scientific Documents ({EEKE2021}) at the {ACM/IEEE} Joint
  Conference on Digital Libraries 2021 ({JCDL2021})}, Online.

\bibitem[{Vanetik et~al.(2020)Vanetik, Litvak, Shevchuk, and
  Reznik}]{vanetik-etal-2020-automated}
Natalia Vanetik, Marina Litvak, Sergey Shevchuk, and Lior Reznik. 2020.
\newblock \href {https://aclanthology.org/2020.lrec-1.256} {Automated discovery
  of mathematical definitions in text}.
\newblock In \emph{Proceedings of the Twelfth Language Resources and Evaluation
  Conference}, pages 2086--2094, Marseille, France. European Language Resources
  Association.

\bibitem[{Veyseh et~al.(2019)Veyseh, Dernoncourt, Dou, and
  Nguyen}]{ben_veyseh_joint_2019}
Amir Pouran~Ben Veyseh, Franck Dernoncourt, Dejing Dou, and Thien~Huu Nguyen.
  2019.
\newblock \href {https://doi.org/10.48550/arXiv.1911.01678} {A {Joint} {Model}
  for {Definition} {Extraction} with {Syntactic} {Connection} and {Semantic}
  {Consistency}}.

\bibitem[{Wadden et~al.(2019)Wadden, Wennberg, Luan, and
  Hajishirzi}]{Wadden2019EntityRA}
David Wadden, Ulme Wennberg, Yi~Luan, and Hannaneh Hajishirzi. 2019.
\newblock Entity, relation, and event extraction with contextualized span
  representations.
\newblock \emph{ArXiv}, abs/1909.03546.

\bibitem[{Ye et~al.(2022)Ye, Lin, Li, and Sun}]{ye2022plmarker}
Deming Ye, Yankai Lin, Peng Li, and Maosong Sun. 2022.
\newblock \href {https://aclanthology.org/2022.acl-long.337} {Packed levitated
  marker for entity and relation extraction}.
\newblock In \emph{Proceedings of the 60th Annual Meeting of the Association
  for Computational Linguistics (Volume 1: Long Papers), {ACL} 2022, Dublin,
  Ireland, May 22-27, 2022}, pages 4904--4917. Association for Computational
  Linguistics.

\end{thebibliography}

\section{Language Resource References}
\label{lr:ref}
\bibliographystylelanguageresource{lrec-coling2024-natbib}
\bibliographylanguageresource{languageresource}

\section{Code Availability}

All of the code necessary to reproduce the experiments, build the corpus, and run the interface are freely available at \url{https://github.com/ToposInstitute/parmesan_benchmarks} and \url{https://github.com/ToposInstitute/parmesan}.

\end{document}